\documentclass[letterpaper, 10 pt, conference]{ieeeconf}  % Comment this line out if you need a4paper

\IEEEoverridecommandlockouts                              % This command is only needed if 
\usepackage[sorting=none, style=numeric-comp, maxbibnames=99]{biblatex}

\usepackage{color}
\usepackage{epsfig}
\usepackage{graphicx}
\usepackage{algorithm,algorithmic}
\usepackage{adjustbox}
\usepackage{array}
\usepackage{booktabs}
\usepackage{colortbl}
\usepackage{float,wrapfig}
\usepackage{framed}
\usepackage{hhline}
\usepackage{multirow}
\usepackage{subcaption} % issues a warning with CVPR/ICCV format
\usepackage[font=small]{caption}
\usepackage[percent]{overpic}
\usepackage{amsmath,amsfonts,amssymb}
\usepackage{amsthm} 
\usepackage{bm}
\usepackage{nicefrac}
\usepackage{microtype}
\usepackage{contour}
\usepackage{courier}
\usepackage{changepage}
\usepackage{extramarks}
\usepackage{fancyhdr}
\usepackage{lastpage}
\usepackage{setspace}
\usepackage{soul}
\usepackage{xspace}
\usepackage{cuted}
\usepackage{fancybox}
\usepackage{afterpage}
\usepackage[breaklinks=true,colorlinks,backref=True]{hyperref}
\hypersetup{colorlinks,linkcolor={black},citecolor={MSBlue},urlcolor={magenta}}
\usepackage{url}
\usepackage{quoting}
\usepackage{epigraph}

\usepackage{enumerate}
\usepackage{paralist,tabularx}
\usepackage{comment}
\usepackage{pdfpages}
\usepackage{caption}
\usepackage{subcaption}

\usepackage{pifont}% http://ctan.org/pkg/pifont

\usepackage{MnSymbol}

\usepackage{enumitem}
\makeatletter
\DeclareRobustCommand\onedot{\futurelet\@let@token\@onedot}
\def\@onedot{\ifx\@let@token.\else.\null\fi\xspace}

\def\etal{et al\onedot}

\makeatother

\definecolor{MyDarkBlue}{rgb}{0,0.08,1}
\definecolor{MyDarkGreen}{rgb}{0.02,0.6,0.02}
\definecolor{MyDarkRed}{rgb}{0.8,0.02,0.02}
\definecolor{MyDarkOrange}{rgb}{0.40,0.2,0.02}
\definecolor{MyPurple}{RGB}{111,0,255}
\definecolor{MyRed}{rgb}{1.0,0.0,0.0}
\definecolor{MyGold}{rgb}{0.75,0.6,0.12}
\definecolor{MyDarkgray}{rgb}{0.66, 0.66, 0.66}
\definecolor{MyPink}{rgb}{1, 0.75, 0.79}
\definecolor{GreenStarColor}{rgb}{0.54, 0.84, 0.41}
\definecolor{MSBlue}{rgb}{0, 0.35, 0.49}

\DeclareMathOperator*{\argmin}{argmin} % thin space, limits underneath in displays

\title{\LARGE \bf Task-Based Design and Policy Co-Optimization\\for Tendon-driven Underactuated Kinematic Chains}

\author{
Sharfin Islam\authorrefmark{1}\authorrefmark{2}, %
Zhanpeng He\authorrefmark{1}\authorrefmark{3}, %
Matei Ciocarlie\authorrefmark{2} \\%
\url{https://roamlab.github.io/tentamorph/}
\thanks{
\authorrefmark{1}Equal contributions.
\authorrefmark{2}Dept. of Mechanical Engineering, \authorrefmark{3}Dept. of Computer Science, Columbia University, New York, USA}
}
 \AtBeginBibliography{\small}

\begin{document}

\maketitle
\thispagestyle{empty}
\pagestyle{empty}

%%%%%%%%%%%%%%%%%%%%%%%%%%%%%%%%%%%%%%%%%%%%%%%%%%%%%%%%%%%%%%%%%%%%%%%%%%%%%%%%
\begin{abstract}

Underactuated manipulators reduce the number of bulky motors, thereby enabling compact and mechanically robust designs. However, fewer actuators than joints means that the manipulator can only access a specific manifold within the joint space, which is particular to a given hardware configuration and can be low-dimensional and/or discontinuous. Determining an appropriate set of hardware parameters for this class of mechanisms, therefore, is difficult - even for traditional task-based co-optimization methods.  In this paper, our goal is to implement a task-based design and policy co-optimization method for underactuated, tendon-driven manipulators. We first formulate a general model for an underactuated, tendon-driven transmission. We then use this model to co-optimize a three-link, two-actuator kinematic chain using reinforcement learning. We demonstrate that our optimized tendon transmission and control policy can be transferred reliably to physical hardware with real-world reaching experiments. 
% Although our experiments are limited to single kinematic chain and only reaching tasks, our overall goal is to broaden the set of tools available to exploit best the benefits posed by underactuated, tendon-driven manipulators.  

\end{abstract}

%%%%%%%%%%%%%%%%%%%%%%%%%%%%%%%%%%%%%%%%%%%%%%%%%%%%%%%%%%%%%%%%%%%%%%%%%%%%%%%%
\section{Introduction}
\label{sec:intro}

Underactuated manipulators require a carefully designed transmission, often tendon-driven, to take advantage of a reduced number of actuators in the robot. Such designs range from planar serial chains with relatively few links to complex, hyper-redundant continuum robots \cite{HIROSE1978351,matei2010data,matei2013kinetic,rojas2021force,jbk2021frontiers,continuumReview,yi2023simultaneous}. In all of these cases, being able to reduce the number of actuators means that we can build smaller and more lightweight designs, place actuation at more proximal locations in the chain, and take advantage of passive compliance in the un-actuated degrees of freedom.

However, the compromise of such designs is that, with fewer actuators than degrees of freedom, underactuated manipulators can directly access only a certain manifold within the overall state space. This manifold, which contains the set of obtainable states for a particular hardware configuration, can be low dimensional and/or discontinuous. These limitations affect our ability to plan controllers that smoothly transition between various states to accomplish a desired task. The process of tuning the hardware parameters in order to ensure the set of accessible states matches the desired task can be slow and cumbersome, particularly if, as is often the case, changes in the design parameters of the underactuated transmission have a counter-intuitive effect the overall behavior of the robot.

\begin{figure}
    \centering
    \includegraphics[width=0.98\linewidth]{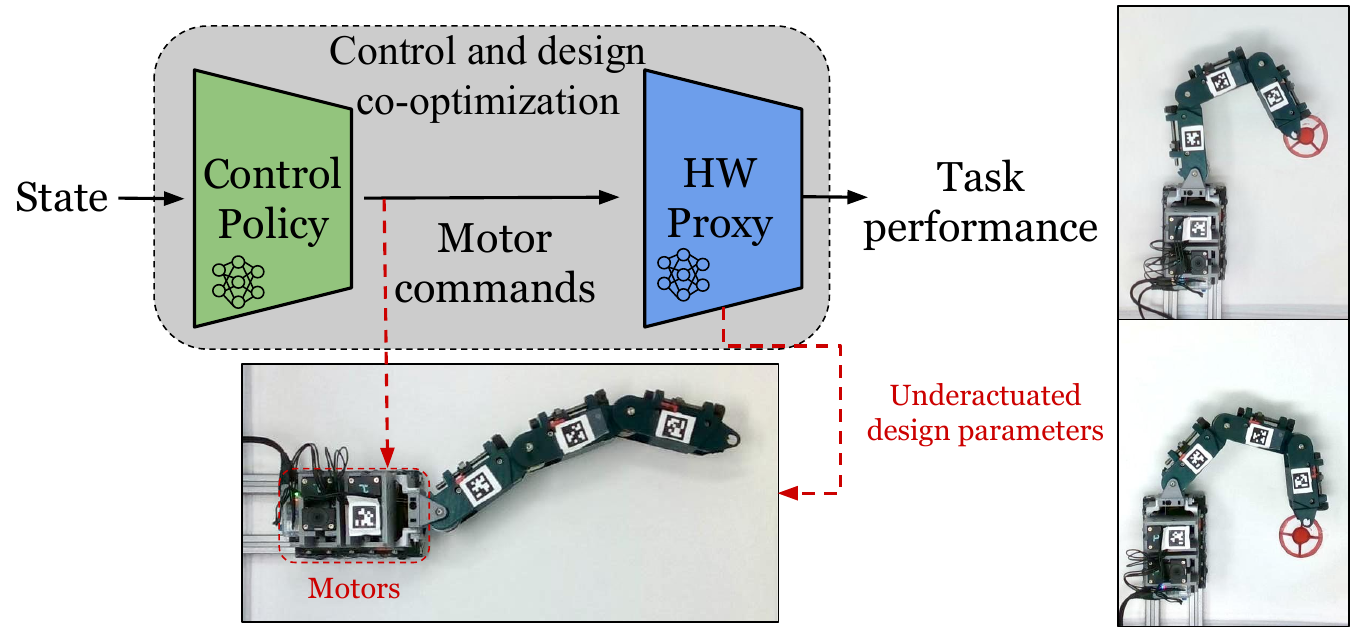}
    \caption{We optimize an underactuated, tendon-driven transmission for kinematic chains. We formulate and parameterize a general model for N links and M actuators. We apply our model to a three-link, two-actuator tentacle that we co-optimize using reinforcement learning (top row, left). We then validate our results on physical hardware (right).}
    \label{fig:eye-candy}
    \vspace{-2em}
\end{figure}

Co-optimization, or the process of simultaneously searching the space of both hardware and control, is a possible solution to the problem of ensuring that a given hardware design is capable of a desired task. Such methods have been attempted in the context of underactuated kinematic chains, but often restricted to simulation \cite{ma2013,Deimel2017AutomatedCO,Barbazza2017}, limited validation of the simulated controller on real hardware \cite{Xu2021AnED}, or only implemented on single-actuator systems \cite{chen2020hwasp}. The fundamental difficulty of such approaches lies in formulating a co-design problem that a) enables the use of non-trivial controllers or policies, b) can be solved to an acceptable optimum point, c) guarantees that the final result can be physically realized in real hardware, and finally d) ensures that the optimized control policy also transfers to real hardware without substantial loss of performance. This is a difficult set of goals to achieve simultaneously, and, to the best of our knowledge, no current method has done so for underactuated, tendon-driven transmissions with multiple actuators.

In this paper, we start with an underactuated tendon transmission model for general planar kinematic chains, which can capture a diverse set of hardware configurations. We use a forward model that captures the behavior of the system as a function of design parameters and control inputs. We then show that this model enables end-to-end co-optimization of design and control policies for a specified task. To solve the co-optimization problem, we adopt MORPH \cite{He2023MORPH}, a recently introduced end-to-end co-optimization framework that uses a proxy model to mimic the effect of hardware, and generate gradients of hardware parameters w.r.t. tasks performance. This allows us to learn both control and hardware parameters to accomplish desired tasks. We then show that it is possible to build a real robotic manipulator with this optimized transmission, and validate that our optimized control policy can be transferred reliably to real hardware. 

An additional feature of our hardware design is that some design parameters can be modified without requiring re-assembly of the manipulator, and by using the same set of fabricated components. This allows for different levels of flexibility in robot behaviors. If the task requires explorations of diverse behaviors, we can optimize all hardware parameters to explore a large solution space. In the case that the task is simpler, we can optimize only the easily adaptable parameters to avoid reassembly. While in this paper we focus on a single kinematic chain optimized for simple reaching tasks, our directional goal to enable underactuated, tendon transmission models that can be co-optimized along with their control policies, which can in turn transfer to the real world. Overall, the main contributions of this paper include:
\begin{itemize}
    \item We formulate a model of underactuated, tendon-driven, passively-compliant planar kinematic chains that enables efficient end-to-end task-based optimization of the design and control parameters. 
    \item We show that the results of the co-optimization process can be transferred to real hardware implementing the optimized design parameters. The resulting robots can then run the optimized control policy which also achieves sim2real transfer in task performance.
    \item To the best of our knowledge, this is the first time that task-based, policy and design co-optimization methods have been demonstrated for underactuated manipulators with multi-dimensional manifolds. 
\end{itemize}

\section{Related Work}

An underactuated manipulator is a mechanical system composed of links and joints that has fewer actuators than degrees of freedom \cite{Jiang2011}. The transmission of these manipulators are often tendon-driven, as a single actuated tendon can be routed to control multiple joints. Reducing the number of bulky actuators enables designers to build more compact and lightweight manipulators. Moreover, tendon-transmission allow designers to dislocate the motors from the joints. Dislocating the actuators reduces the inertia of the links and makes it easier to make the manipulator robust to water, dust, and other difficult environmental conditions \cite{ozawa2009design}. With so many practical benefits, a myriad of diverse underactuated, tendon-driven manipulators have been proposed over the past several decades.

There is an extensive history of underactuated manipulator design. The first underactuated manipulator, Hirose's soft gripper introduced in 1978, was designed to softly capture a diverse range of objects with uniform pressure \cite{HIROSE1978351}. Over time, these manipulator designs became much more advanced and their applications now extend to more robust and intricate grasping behaviors \cite{dollar2009,su2012,Grioli2012,santina2018,chen2021}. Currently, there is extensive research in the design and control of underactuated manipulators for tasks even more dexterous than grasping, such as in-hand manipulation \cite{Udawatta2003, odhner2011,morgan2022}.To realize these more advanced capabilities, the tendon-transmission of these manipulators had to be carefully designed, optimized, and further hand-tuned. This process is cumbersome and time-consuming, but necessary. Co-optimization methods are a possible tool to help design these complicated mechanisms. With recent advancements in reinforcement learning, co-optimization is now more powerful than ever.

Recently, researchers have considered reinforcement learning for task-driven design and control co-optimization ~\cite{jackson2021orchid, Ha2018ReinforcementLF, schaff2019jointly, wang2018neural}. The key of this line of work is to derive policy gradient w.r.t both the control and design parameters. Chen and He \etal \cite{chen2020hwasp} propose to integrate a differentiable model of the hardware with a control policy to adapt hardware design via policy gradients. A key limitation of this approach is the requirement of differentiable physics simulation. In cases when the forward transition cannot be modeled in a differentiable manner, researchers propose methods to learn design parameters in the input or output space of a policy. For instance, Luck \etal \cite{Luck2019DataefficientCO} propose to learn an expressive latent space to represent the design parameters and condition a policy with a latent design representation.
Transform2Act \cite{Yuan2021Transform2ActLA} propose to have a transform stage in their policy that estimates actions to modify a robot's design and a control stage that computes control sequences given a specific design. 
Our hardware design is not differentiable. Hence, in this work, we apply MORPH \cite{He2023MORPH}, a method that co-optimizes design and control in parameter space that does not require differentiable physics.

\section{Method}
\label{sec:method}

We formalize our problem as follows. Our goal is to build a kinematic chain (i.e., tentacle/trunk) robot that can achieve flexible behaviors, such as reaching desired parts of the workspace, while maintaining the practical design benefits of underactuation. For this class of mechanisms, the design parameters and controller for a given task are innately coupled. Therefore, we take a task-based hardware optimization approach to search for hardware parameters that yield manifolds such that we can smoothly transition between desired states. 

Our method has three main components: 1) an under-actuated, tendon-driven transmission design for compliant kinematic chains; 2) a model that captures the forward dynamics of our designed robot;
and 3) an end-to-end co-optimization pipeline that can directly learn the parameters of the proposed design along with a control policy for a given reaching task. 

\subsection{Transmission design}

\begin{figure}
    \centering
    \includegraphics[width=\linewidth]{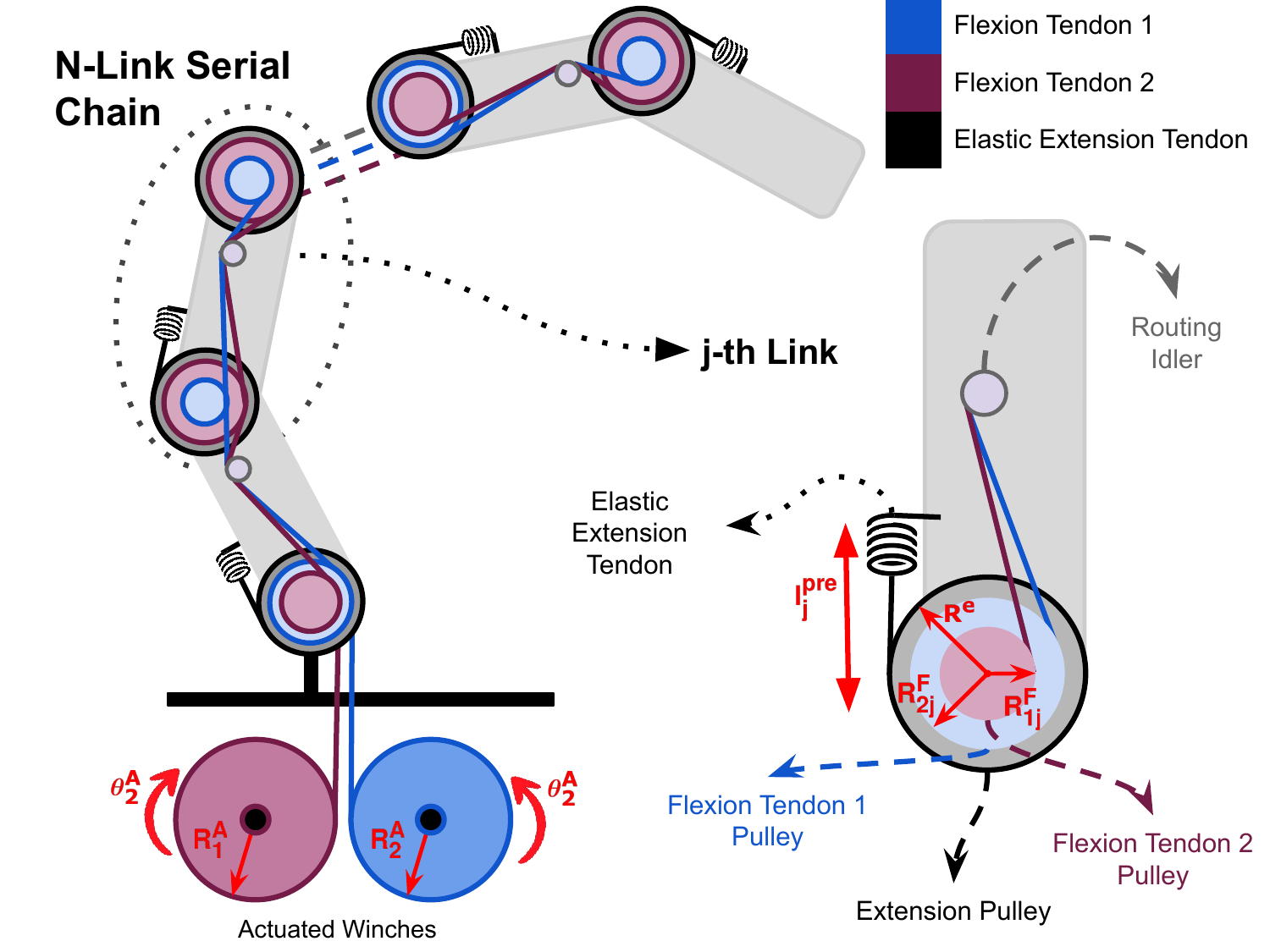}
     \caption{Our flexible tendon-transmission design for compliant, underactuated kinematic chains. In this design, $N$ links are driven by $M$ actuated flexion tendons. We also implement a passive extension mechanism, which we can precisely pre-tension. The parameters of our transmission are all flexion tendon radii, extension tendon radii, and elastic tendon pre-elongations.} 
    \label{fig:tendon_model}
    \vspace{-2em}
\end{figure}

Our design, shown in Figs. \ref{fig:eye-candy} and \ref{fig:tendon_model}, assume a kinematic chain with $N$ links driven by $M$ motors. All motors are located inside the base and actuate winches that are connected to the joints via tendons. For each motor $i$, we denote the radius of the actuator as $R^A_i$. We assume that each motor is driving a tendon that traverses the entire length of the chain, thereby helping flex every joint. We denote the length of each link $j$ as $L_j$. For a the corresponding joint $j$, each motor $i$ will wrap around a pulley. We use ${R_{ij}}^f$ to denote the radius of the flexion pulley for this joint and motor pair. 

We assume all actuators provide flexion forces, and the transmission uses purely passive extension mechanisms. At each joint, the mechanism features a passive elastic tendon that stretches over a pulley of constant radius (denoted by $R^e$) to provide a restoring extension torque. Each elastic tendon can be pretensioned individually; we use ${l_{j}}^{pre}$ to denote the pretensioning elongation of the passive tendon at joint $j$.

This mechanism has a number of desirable characteristics. Underactuation leads to a small number of motors, and placing all motors inside the base frees the links of the kinematic chain from any motors or electronics. However, the movement of the robot is non-trivial to define or control. Critically, robot movement is determined not only by the actuators but also by a number of design parameters. These include all flexion tendon radii, extension tendon radii, and elastic tendon pre-elongations. Furthermore, the ability of a robot to reach specific points in its workspace is clearly also determined by the lengths of the links. Our goal is to devise a procedure capable of optimizing all these design parameters while providing a policy for controlling motor movements in order to achieve a specific task. 

In addition, our proposed design makes some of the design parameters easier to change than others. In particular, we mount each elastic tendon on a linear slider mechanism. The position of this slider is set by rotating a lead-screw, thereby controlling the pre-tension elongations of the elastic extension tendons to less than 1mm of precision. This means that some of our design parameters (pulley radii, link lengths) are more difficult to change, as they require a full reassembly, while others (elastic tendon elongations) are easier to change depending on the task. We want to leverage this ability in our co-optimization procedure.

\subsection{Forward actuation model for our transmission design}

In order to enable a co-optimization routine for our transmission design, we first need a forward actuation model that relates the hardware parameters and the actuator inputs to the resulting state of the manipulator. The input to this model consists of the servo angle of the winches that control flexion tendons, which we write as a column vector $\theta_A$ of size $\mathbb{R}^{M \times 1}$. The next state is defined by the set of joint angles $\theta_J$ of size $\mathbb{R}^{N \times 1}$. In other words, the actuation model must predict joint angles $\theta_J$ as a function of the actuator commands $\theta_A$, as well as all design parameters described in the previous section. We formulate this actuation model by searching for  $\theta_J$ that minimizes the total stored energy, denoted by  $U$, while still meeting the constraints imposed by the rigid flexion tendons.

We begin our model by first formulating this constraint: since the flexion tendons cannot extend, we know that the tendons will either be in tension or accumulate slack. This slack is a difference between the collective change in length of the tendon due to the motion of the joints versus the change in length due to servo-driven winch. The slack, therefore, is a function of the flexion pulley radii, actuated winch radii, and joint angles. 

The flexion radii can be composed into a matrix of size $\mathbb{R}^{M \times N} $ as follows: 

\begin{align}
    \label{rFlexMat}
    \mathbf{R^F} = \begin{bmatrix}
        R_{11}^F & R_{12}^F & ... & R_{1N}^F \\
        R_{21}^F & R_{22}^F & ... & R_{2N}^F \\
        ... &...&...&... \\
        R_{M1}^F & R_{M2}^F & ... & R_{MN}^F \\
    \end{bmatrix}  
\end{align}

The radii of the actuated winches can be similarly composed into a diagonal matrix of size $\mathbb{R}^{M \times M}$.

\begin{align}
    \label{rAmat}
    \mathbf{R^A} = \begin{bmatrix}
        R_{1}^A & 0 & ... & 0 \\
        0 & R_{2}^A & ... & 0 \\
        ... &...&...&... \\
        0 & 0 & ... & R_{M}^A \\
    \end{bmatrix}  
\end{align}

With these matrices, we can now define a column vector of the slack collected in the tendons $(w)$. 

\begin{align}
    \label{total_elongation}
     w = \mathbf{R^F} \theta_J - \mathbf{R^A}\theta_A
\end{align}

\vspace{1em}

The slack is important as it limits the set of obtainable states for any given action. We know that the robot will settle at a configuration in this set of states that will minimize the overall stored elastic energy $(U)$ \cite{Jiang2011}. This stored energy comes from the preloaded elongation $l_{pre}$ of the elastic tendons, but also the elongation due to joint flexion. The total elongation for any j-th link $(l_j)$ can be written as follows:
\vspace{-.05em}
\begin{align}
    \label{total_elongation}
    \Delta l_j= R^e \theta_{j} + l^{pre}_j
\end{align}

\label{sec:forward_dynamics}
We can now pose the forward actuation model to solve for the next state as $ \theta_J' = f(\theta_J,  \theta_A)$ relating the state to the action as a numerical optimization problem: given $\theta_A$,$\theta_J$,
\begin{align}
    % \text{Given } \vec{\theta_A}, &\vec{\theta_J}, \nonumber\\
    % Find  } \vec{\Delta \theta_J} \nonumber\\
    \argmin_{\theta_J} & U=\frac{1}{2}\sum_{j=1}^N k l_j^2 \label{storedEnergy}\\
    \text{subject to } &\text{\hspace{1em} } w \geq 0 \label{const1} \\
    & \text{\hspace{1em} } det(diag(w)) = 0  \label{const2}
\end{align}

\begin{figure}
    \centering
    \includegraphics[width=\linewidth]{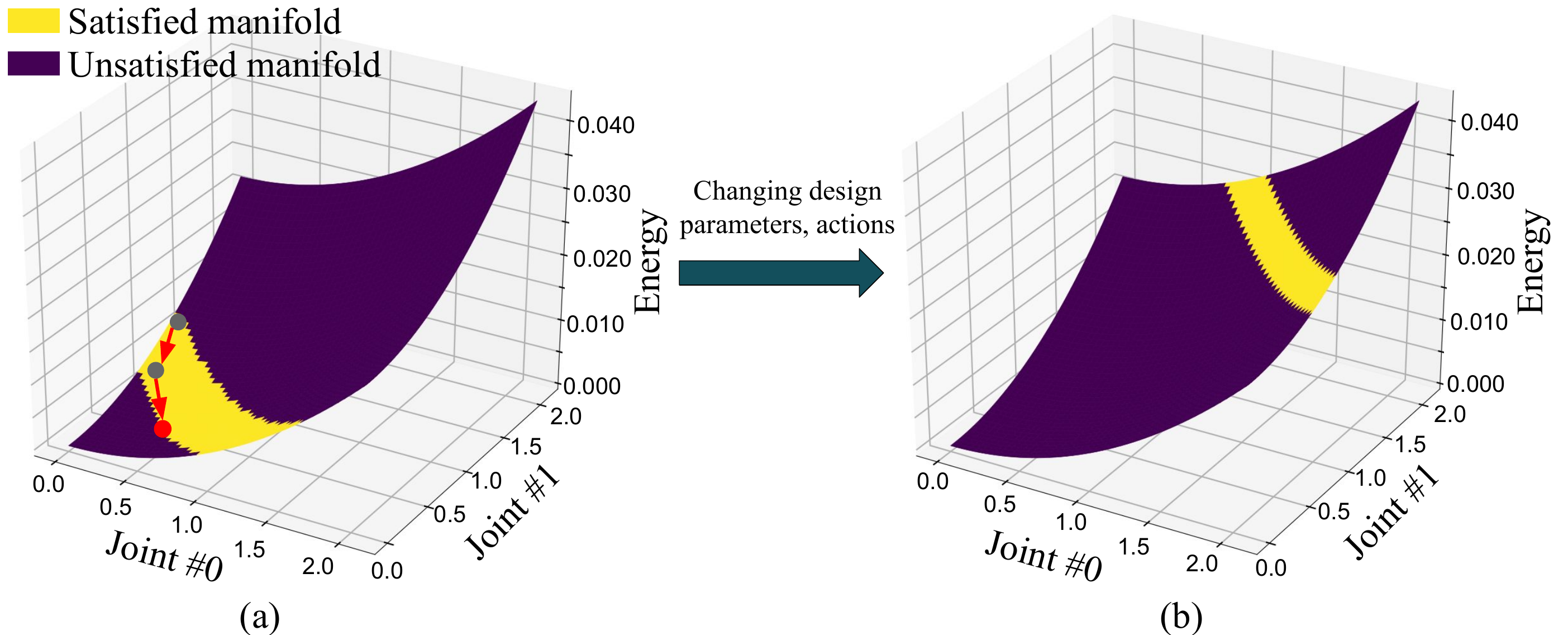}
     \caption{\textbf{Illustrations of our optimization-based forward actuation model.} (a) and (b) show the global energy landscape (z-axis). The yellow regions are manifolds that satisfy our constraints (Eq.6 and Eq.7). The red arrows in (a) represent optimization steps (Eq.5) that find the energy minimum inside the manifold. Changing design parameters and control actions result in a change in the manifold.} 
    \label{fig:energy-manifold}
\end{figure}

Since the flexion tendons are in-extensible, the change in length of the tendon at the joints must always be greater than the change in length of the tendon due to the actuator. The slack, therefore, must always be greater than zero, as shown in Eq. \ref{const1}. Additionally, if the joint angles are non-zero, the tendon must always be in tension, and therefore, at least one element in $\vec{w}$ must be zero. This constraint is shown in Eq. \ref{const2}. We enforce these constraints While minimizing the total stored energy given in Eq.\ref{storedEnergy},

Actions and hardware parameters do not change the global energy landscape, but instead, the section of the landscape that satisfies our constraints (see Fig.\ref{fig:energy-manifold}). 
Hence, optimizing control $\theta_A$ and hardware parameters (e.g., $\mathbf{R_F}$ and $R_e$) is finding appropriate energy manifolds whose minimum energy states are ones we want to visit for task completion. 
There exists some combination of hardware parameters that ensures these manifolds are continuous are have a clear energy minima. Arriving at this specific set of parameters relies on observing how the changes in parameters affect the most suitable control strategy for solving a given task.
Finally, in this work, we focus on optimizing the following set of hardware parameters: $\phi = (\mathbf{R^F}, l_{pre}, L)$.

\subsection{Task-aware co-optimization of design and control}

Equipped with the forward actuation model described in the previous section, we can now focus on the problem of co-optimizing a set of design parameters and a motor control policy for a given task. Using a standard reinforcement learning (RL) formalism, we model each task as a Markov Decision Process (MDP), or a tuple $(\mathcal{S}, \mathcal{A}, \mathcal{F}, \mathcal{R})$, where $\mathcal{S}$ is the state space, $\mathcal{A}$ is the action space, $\mathcal{R}(\bm{s}, \bm{a})$ is the reward function, and $\bm{s}' = \mathcal{F}( \bm{s}, \bm{a})$ is the forward transition model, where $\bm{s}, \bm{s}' \in \mathcal{S}$, and $\bm{a} \in \mathcal{A}$. 

In this work, we use a state vector $s$ containing the joint angles $\theta_J$, motor positions $\theta_A$, and 2D end-effector positions $\mathrm{x}_{e.e.}$. Actions for a control policy are motor commands $\theta_A$. The reward function $\mathcal{R}(\bm{s}, \bm{a})$ encodes the desired performance; for example, if we want the end-effector to reach a specific point, the reward will comprise a negative distance to the goal. For each task, we want to find the parameters of a control policy $\pi_\theta$ as well as the design parameters $\phi$ that optimize task performance, which is evaluated by its expected returns: $\mathbb{E}[\sum_{t=0}^T\mathcal{R}(\bm{s}_t,\bm{a}_t)]$. 

The state transition model $\mathcal{F}$ requires additional considerations. 
As described in Section \ref{sec:forward_dynamics}, it consists of an optimization-based model $f$ whose behaviors are determined by some of the hardware parameters from $\phi$. However, this model is non-differentiable and, therefore, can not be used in a standard policy gradient optimization routine.

To co-optimize the hardware design with control, we apply MORPH~\cite{He2023MORPH}, an end-to-end co-optimization method that uses a neural network proxy model $h^{nn}$ to approximate the forward transition $\mathcal{F}$. 
The proxy model and a control policy are co-optimized with task performance while asking $h^{nn}$ to be close to $\mathcal{F}$. 
In this work, instead of mimicking just the forward transition, we consider the effect of hardware design parameters in task space. 
Hence, we ask the hardware proxy to approximate both the forward transition and forward kinematics: $q_{\phi} = f \circ T_{FK}$. Note that our hardware parameters $\phi$ are encapsulated in different parts of $q$. For example, link lengths only affect forward kinematics, while pulley radii and preloaded tension govern the behaviors of $f$.
Hence, the overall optimization objective is:
% As defined by He \etal, the combined policy have two components: a control policy $\pi$ that takes observations $o$ as inputs and infers motor commands $a_{motor}$ and a hardware proxy model $h_{nn}$ that estimate task space action from observations and motor commands. While being co-optimized to task performance, $h_{nn}$ is constrained to be close to the hardware model. Hence, the overall optimization objective has two components: 
\begin{align}
    \label{eq:objective}
    &\max_{\theta, \psi}\mathbb{E}_{\pi, h^{nn}}[\underbrace{\sum_{t=0}^T\mathcal{R}(s_t, \theta_{A})}_{\text{task performance}} + \underbrace{\alpha ||h^{nn}(\theta_J,\theta_A)- q(\theta_J, \theta_A)||^2}_{\text{hardware constraint}}]
\end{align}
where $\alpha$ is a constant number.

To derive explicit hardware design parameters, for every $N$ epochs, we also use CMA-ES \cite{hansen2001completely} to search $\phi$ to match the updated hardware proxy $h^{nn}$:
\begin{align}
    \label{eq:obj2}
    &\min_{\phi}||h^{nn}(\theta_J, \theta_A)- q_{\phi}(\theta_J, \theta_A)||^2
\end{align}

The final algorithm is an iterative process: We first co-optimize the hardware proxy and the control policy to improve task performance, then extract explicit hardware parameters that match the current version of the hardware proxy.

\section{Experimental Set-up}
\label{sec:experiments}
\subsection{Design and control co-optimization}
To evaluate our method, we optimize the aforementioned design with several goal-reaching tasks, i.e., reaching a goal location in 2D with its end-effector. 
We have three sets of experiments demonstrating three ways to utilize our co-optimization pipeline for flexible motor skills: 
\begin{enumerate}

    \item \textbf{Goal reaching via re-fabrication}: We adapt all design and control parameters to different goals but only train each policy for a single goal. In this specific experiment, reaching multiple goals requires the re-fabrication of a real robot since the link length and pulley radii cannot be changed after assembly without fabricating new parts and reassembling the manipulator. 
    \item \textbf{Goal reaching via online hardware updates}: In this experiment, we optimize the hardware design in two stages: we first optimize design parameters that require re-fabrication (i.e., pulley radii) to a specific goal. Then, we fix all parameters except the pre-tension elongation and co-optimize for a different goal. 
    \item \textbf{Goal reaching via multi-goal control policies}: We optimize all parameters to learn a shared design and control that can achieve multiple goals with a single control policy. In this case, we do not need to adapt any of the hardware parameters to reach multiple goals after the policy is trained. 
\end{enumerate}

The first experiment is our design's most limited application due to the need for re-fabrication. However, it demonstrates that our transmission can be optimized to reach different areas of the fully actuated workspace. We also train control policies with fixed initial design parameters to compare task performance. Fig.\ref{fig:reachers} shows the goal distribution for the first experiment set.
% In the simulation, we validated that the goal in this experiment could not be reached with the initial value of the design parameters. 
% Our co-optimization method, in this experiment, shows that our robot design is flexible. 
% With our transmission model, we can co-optimize the design and control to reach different areas of the full robot workspace with a stable action sequence. 
The second set of experiments leverages a key aspect of our design - the ability to easily adapt some design parameters without having to go through tedious re-design, fabrication, and assembly. 
For this second set, we keep the same pulley radii and only optimize the pretension elongation for a goal different from the first experiment. In the first and second experiments, we use the state space described in Section.\ref{sec:method} for inputs of our control policy.

The last experiment demonstrates that our robot, while being underactuated, can achieve different goals with the same design parameters and control policy—if the desired goals are not too far away. The state space is extended from the first two experiment sets with the addition of the goal locations. To establish the specific set of goals we hope to achieve, We sample goals randomly around a center location $(-0.16, 0)$ within a radius of $50\mathrm{mm}$. For evaluation, we randomly sample $20$ goals and execute the policy $10$ times for each goal. This experiment requires re-assembly of the entire tentacle as we optimize all the design parameters. Therefore, we only run this experiment in simulation. 

For all experiments, We use a dense reward function: $r = ||\mathrm{x}_{e.e.} - \mathrm{x}_{goal}||^2 + b$, where $\mathrm{x}_{e.e}$ and $\mathrm{x}_{goal}$ are the position of the end-effector and the goal, $b$ is a task completion bonus that is provided when the distance of E.E. and the goal is less than $2\mathrm{mm}$. We initialize the hardware parameters to be $R^{F}_1 = [0.005, 0.005, 0.005]$ and $R^{F}_2=[0.005, 0.01, 0.02]$, $L = [0.8, 0.8, 0.8]$ and $l^{pre} = [0, 0, 0]$. The units of all these parameters is meters. 

\begin{figure}
    \centering
    \includegraphics[width=0.8\linewidth]{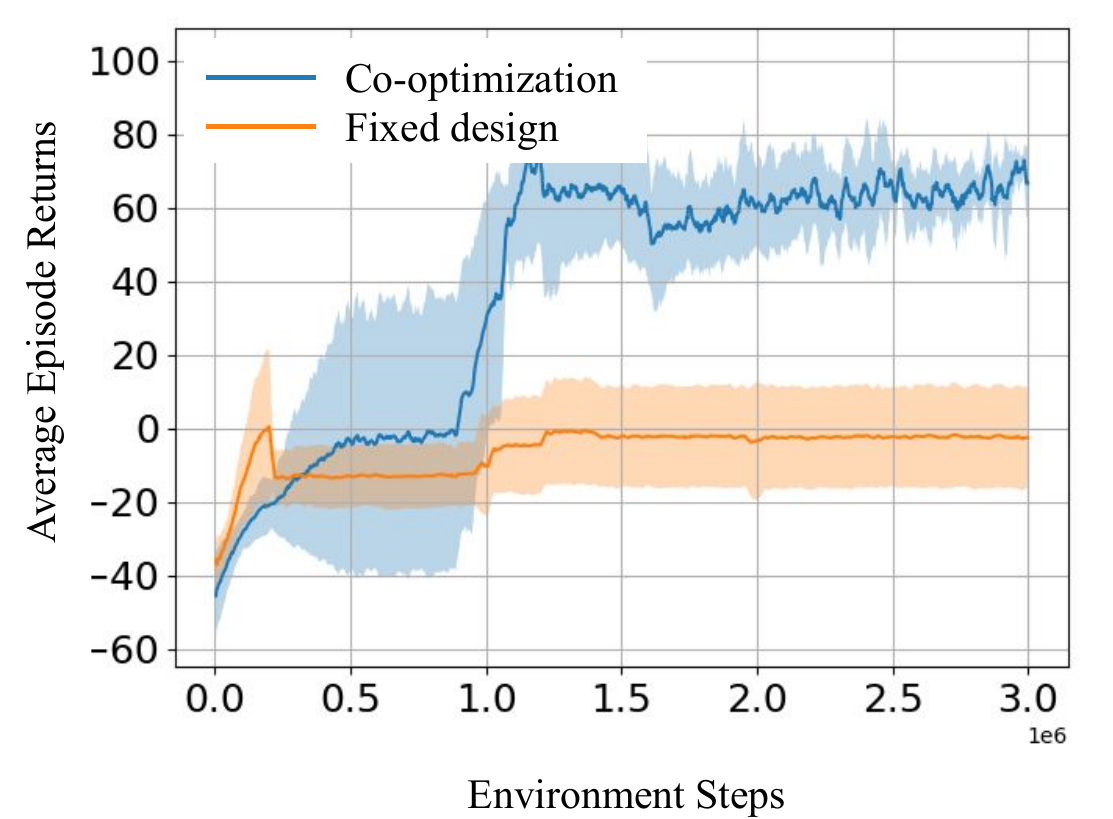}
    \caption{Average episode returns for goal reaching via re-fabrication.}
    \label{fig:rewards}
\end{figure}
\subsection{Hardware implementation and sim-to-real transfer}

\begin{figure}[t!]
    \centering
    \includegraphics[width=\linewidth]{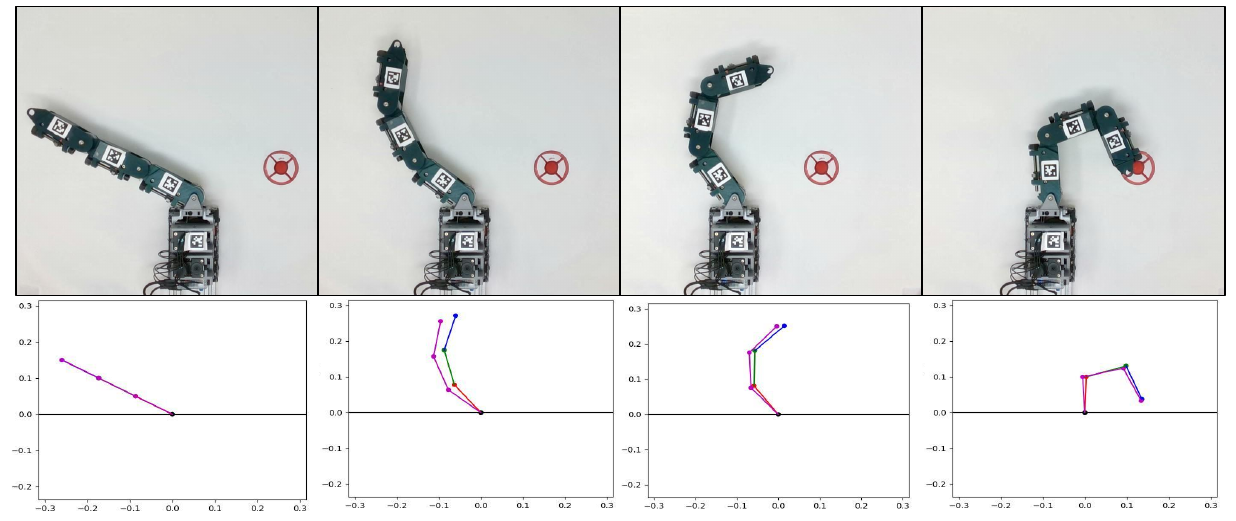}
    \caption{Example of policy rollout on the real robot. Top: real robot rollout. Bottom: comparisons between simulated robots and real robot tracking. The real robot is shown in purple in the bottom row of figures. }
    \label{fig:real-tentacle}
\end{figure}

 Shown in Fig.\ref{fig:eye-candy} and \ref{fig:real-tentacle}, we physically build the manipulator with the optimized pulley radii in the first two experiment sets. To simplify both the fabrication and assembly, we set $(R_e)=15\mathrm{mm}$. Our transmission in this robot is driven by two Dynamixel XM-430 servos that sense and control the angle $(\theta_a)$ of the actuated winches. During our experiments, we fix the manipulator to a experimental rig, which has a mounted camera looking down on the robot. The camera is used in lieu of joint angle encoders, as we calculate the joint angles of our robot $(\theta_J)$by tracking the pose of AR tags attached to our robot as shown in Fig. \ref{fig:real-tentacle}. We use the joint angles collected from the AR tags to execute closed-loop policies for the first two experiment sets. In this case, we close the loop by taking observations from the real robot during policy execution and use our optimized control policy to determine the next action. In addition to closed-loop policies, we also simply train a control policy to reach the goal in simulation, and then directly transfer this policy to the real-robot without adjusting any of the actions during runtime.

 We run both the first and second experiment set on the real robot hardware for both closed-loop and open-loop policies. In the next section, we evaluate the accuracy and precision over 20 samples on the real robot for each set of data.

\section{Results and Analysis}
\label{sec:results}

\subsection{Co-optimization results}

\begin{figure}
    \centering
    \includegraphics[width=\linewidth]{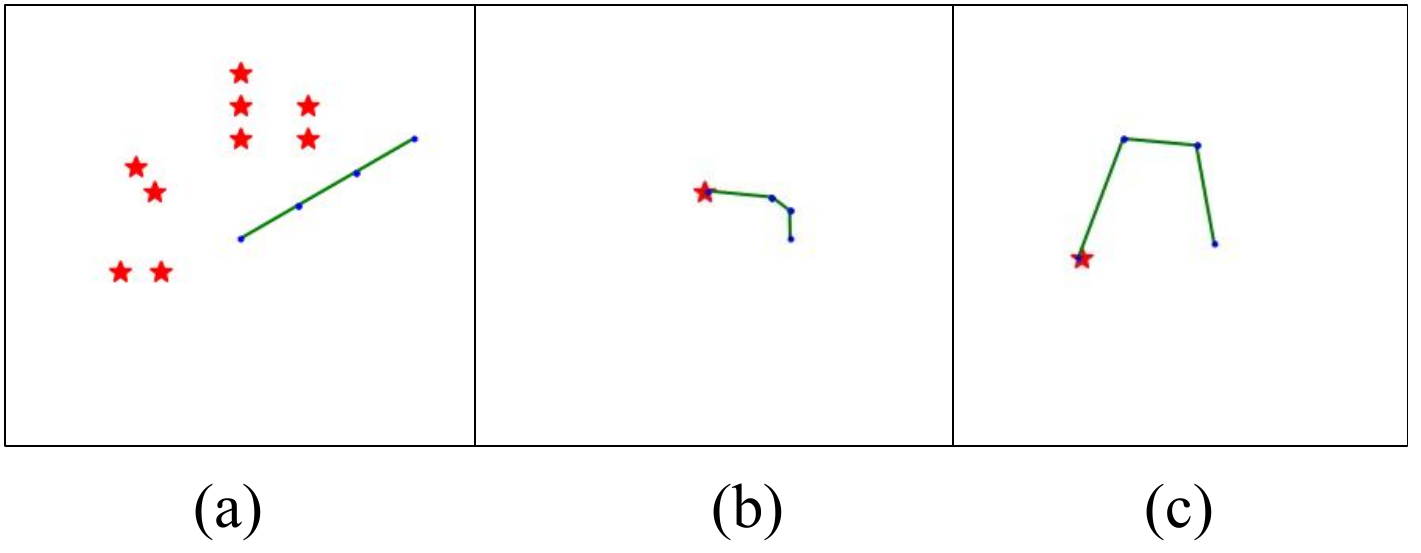}
    \caption{Qualitative results for optimizing underactuated robots to reach different goals in simulation (goal reaching via re-fabrication). (a) shows original design and goal locations. (b) and (c) show two optimized underactuated tentacles reaching different goals. }
    \label{fig:reachers}
\end{figure}

Our results from the first set of experiments show that our model learns different design parameters for different goal locations with an error $1.80\mathrm{mm}$. 
As shown in Fig.\ref{fig:reachers}, our model can adapt hardware parameters to different goals. The initial design can achieve only $1$ out of $9$ goals, achieving much lower average returns than our method (Fig.\ref{fig:rewards}) when we only learn a control policy with a fixed design. 
Although the robot's workspace may contain the goal, the initial hardware parameters make learning a stable control policy difficult. After the design parameters are optimized, the control policy can learn a stable action sequence to arrive at the goal position.

\begin{table}
\centering
    \caption{\textbf{Final Distance to Goal Position of Optimized Robots for real-world reaching tasks.} 
    Open and closed represent open-loop and closed-loop execution, respectively. All units of distance in this table are $\mathrm{mm}$, and each value is calculated for 20 samples.}
    \begin{tabular}{ccccccc}
    \toprule
    & \multicolumn{2}{c}{Avg. distance to goal} & \multicolumn{2}{c}{Std. distance}  \\
    & \multicolumn{2}{c}{\rule{2cm}{0.2pt}} & \multicolumn{2}{c}{\rule{2cm}{0.2pt}} \\
    Stage & Open & Closed & Open & Closed  \\
    \midrule
    Pulley radii  & 13.27 & 6.30 & 7.28 & 2.87  \\
    Preloads & 21.27 & 12.97 & 7.26 & 4.39 \\
    \bottomrule
    \end{tabular}
    \vspace{-1em}
    \label{tab:real results}
\end{table}
\begin{table}
\centering
    \caption{\textbf{Quantitative results for sim-to-real hardware transfer.} Average error between real hardware states and simulated states for both open-loop and closed-loop policy rollouts. Each value is calculated over a total of 40 samples.}
    \begin{tabular}{ccc}
    \toprule
       &  Open-loop & Closed-loop \\
    \midrule
      Joint angle (rad) & 0.176  & 0.0932\\ 
      End-effector position (m) & 17.523  & 10.877\\ 
    \bottomrule
    \end{tabular}
    \label{tab:model accuracy}
\end{table}

% In the second set of experiments, our experiment shows we can achieve another task by adapting only the preload. 
% This allows for fast deployment on real hardware without fabrication. 
% On the real robot, we adjust the extension preload by hand and directly run the control policy. 
% Our results show we can achieve other goals with $35\mathrm{mm}$.

In the second experiment set, our results show we can optimize only the pulley radii to achieve one goal $(0.13, 0.3)$ with $1.8\mathrm{mm}$ error in simulation. Then, we fix the pulley radii and only optimize the preloaded extension to another goal $(0.16, -0.08)$ with $3.2\mathrm{mm}$ error in simulation. Using derived design parameters, we build a real tentacle robot and directly transfer the control policy from simulation to the real world.  

Our results show that the transfer policy achieves, on average, $6.3\mathrm{mm}$ error to the goal location. During training, since the control policy is co-optimized with the non-stationary hardware proxy, it is robust to hardware parameters perturbation. We test both open- and closed-loop performance on the real robot. In the open-loop test case, we record action sequences executed from the simulation and directly execute them on the real robot. The robot can touch the goal with $12.97\mathrm{mm}$ errors.

\subsection{Sim-to-real accuracy}

A key focus in our paper is the ability to transfer the optimized design and control policy to physical robotic hardware.
We achieve accurate transfer in terms of task performance, as shown in Table \ref{tab:real results}. Our closed-loop results show that taking observations directly from the real robot improves task performance substantially.
The closed-loop policies' standard deviation of the distance to the goal is much lower, which indicates that they yield more consistent results in the real world. Additionally, the average final distance to the goal for the closed-loop policies is much lower, indicating higher accuracy. Our policy is robust to actuation errors by reasoning about the current robot state and producing actions that correct its errors. 
In Table \ref{tab:model accuracy}, we calculate the average difference between the real robot state and the simulated robot state for both the closed-loop and open-loop sets of experiments. 
The average error in joint space for closed-loop policy execution is about half the error for open-loop execution. Similarly, the average distance between the real and simulated end-effector positions for the entire action sequence of each experiment is much lower for closed-loop policies.

% \section{Analysis}
\subsection{Energy manifold optimization}

\begin{figure}
    \centering
    \includegraphics[width=\linewidth]{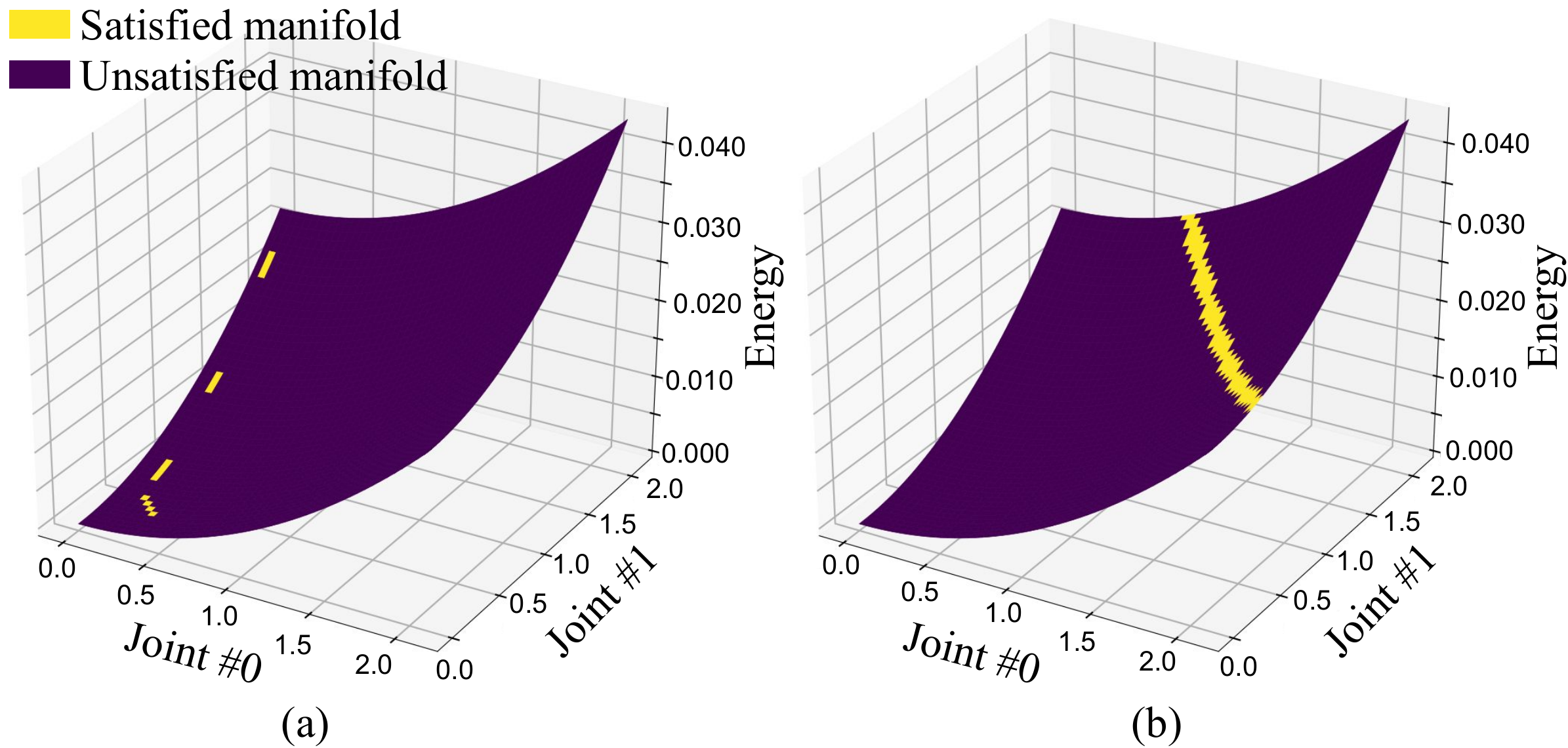}
    \caption{Manifold comparisons before (a) and after (b) hardware optimization. Each grid represents the constraint satisfaction (yellow for constraints satisfied, dark purple for not satisfied) of its center. We set the joint angles of $\#2$ to be 0 for visualization purposes.}
    \label{fig:manifold}
\end{figure}
Our forward model is an optimization-based model. Each forward step requires many optimization steps to find a global energy minimum, which is crucial for accurately simulating our robot. 
Each set of hardware parameters, along with a control action, corresponds to a manifold in the energy landscape. Although the global energy landscape does not have any local optima, a manifold can be discontinuous and has multiple local minimums. This introduces difficulties for optimization and constrains our choice of optimization algorithm to global optimization (e.g., genetic algorithms). However, given the time budget, global optimization can be inefficient and slow down our simulated robot since it generally searches in a large state space.

As mentioned in Section \ref{sec:forward_dynamics}, when we optimize the control and hardware parameters of this robot, we optimize the manifold for energy optimization. 
Our experiments show that our co-optimization process results in an energy manifold that is easy to learn by task. As shown in Fig.\ref{fig:manifold}, the original hardware design parameters provide manifolds that are discontinuous, spread in different regions in the energy landscape, and have several local minimums. 
After MORPH, the resulting manifold (see Fig.\ref{fig:manifold}b) is a continuous and smooth manifold. 
% When a policy select an action, it finds the global energy minimum from this manifold. 
To further analyze the optimized manifold's effect, we apply Sequential Least Squares Programming optimizer (SLSQP) \cite{kraft1988software}, a local optimization algorithm with a time budget of $500$ optimization steps and $100$ random actions, to both the unoptimized and optimized manifolds and compare their results to those of a global search algorithm implemented in-house.
When the hardware is unoptimized, given the same time budget, the simulated results have a much higher error ($0.42$) than the optimized hardware design ($0.18$). This means that the resulting manifold of our optimization is more suitable for fast simulation with local optimization. 

\section{Conclusion}
\label{sec:conclusion}
In this work, we present a method for co-optimizing the design and control of underactuated kinematic chains. The key to our approach is an optimization-based forward actuation model that effectively captures the behavior of our robot design, and a co-optimization pipeline is capable of learning with non-differentiable physics. Our experimental results from simulation and the real world demonstrate that our design, along with the flexibility provided by hardware optimization, results in flexible robot capabilities while enjoying the benefits of underactuation. A key limitation of our current work is the task complexity. While being general, our forward actuation model assumes quasi-static, making contact-rich tasks difficult. In future works, we aim to extend our work to more complex tasks. Another future direction is to further utilize our online adaptable design and explore novel mechanisms to make part of the design adaptable.

\printbibliography

\addtolength{\textheight}{-12cm}   % This command serves to balance the column lengths
\end{document}